\documentclass[11pt]{article}
\usepackage{acl2016}
\usepackage{times}
\usepackage{latexsym}
\usepackage{url}
\usepackage{subfig}

\usepackage{epsfig}
\usepackage{amsmath}
\usepackage{amsfonts}
\usepackage{amssymb}
\usepackage{graphicx}
\usepackage[dvipsnames]{xcolor}

\aclfinalcopy 
\def\aclpaperid{186} 

\newcommand{\tabincell}[2]{\begin{tabular}{@{}#1@{}}#2\end{tabular}}
\newcommand{\enttype}[2]{\underline{\textit{\texttt{#1}}$_{#2}$}}
\DeclareMathOperator\softmax{\textrm{softmax}}

\DeclareMathOperator*{\argmax}{arg\,max}
\DeclareMathOperator*{\minimize}{\textrm{minimize}}

\usepackage{algorithm}
\usepackage[noend]{algpseudocode}

\renewcommand{\algorithmicrequire}{\textbf{Input:}}
\renewcommand{\algorithmicensure}{\textbf{Output:}}
\algnewcommand\algorithmicfuncdesc{\textbf{Function:}}
\algnewcommand\FUNCDESC{\item[\algorithmicfuncdesc]}
\algnewcommand\algorithmicfuncdescb{{\hspace{1.32cm}}}
\algnewcommand\FUNCDESCB{\item[\algorithmicfuncdescb]}
\algnewcommand{\algorithmicgoto}{\textbf{goto}}
\algnewcommand{\Goto}[1]{\algorithmicgoto~\ref{#1}}

\title{Language to Logical Form with Neural Attention}

\author{Li Dong \and Mirella Lapata \\
	Institute for Language, Cognition and Computation \\
	School of Informatics, University of Edinburgh \\
        10 Crichton Street, Edinburgh EH8 9AB \\
	{\tt li.dong@ed.ac.uk}, {\tt mlap@inf.ed.ac.uk}}

\date{}

\begin{document}

\maketitle

\begin{abstract}
Semantic parsing aims at mapping natural language to machine interpretable meaning representations. Traditional approaches rely on high-quality lexicons, manually-built templates, and linguistic features which are either domain- or representation-specific. In this paper we present a general method based on an attention-enhanced encoder-decoder model. We encode input utterances into vector representations, and generate their logical forms by conditioning the output sequences or trees on the encoding vectors. Experimental results on four datasets show that our approach performs competitively without using hand-engineered features and is easy to adapt across domains and meaning representations.
\end{abstract}

\section{Introduction}
\begin{figure}[t!]
	\centering
	\includegraphics[width=0.48\textwidth]{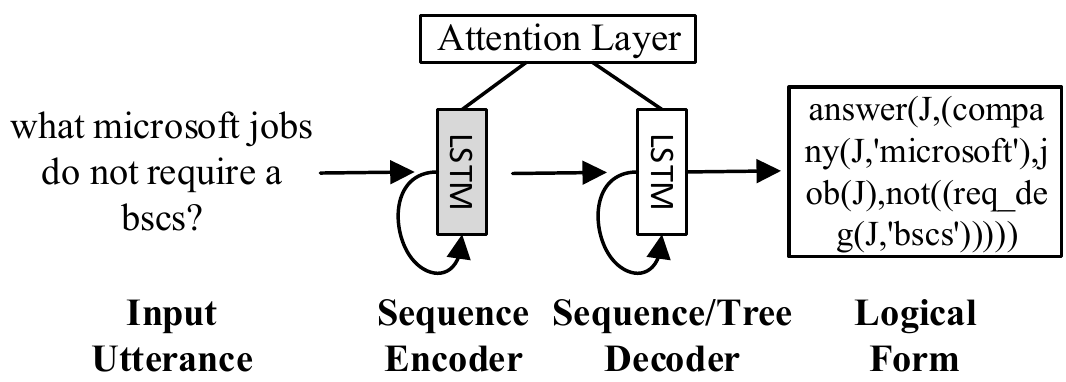}
	\caption{Input utterances and their logical forms are encoded
          and decoded with neural networks. An attention layer is used
          to learn soft alignments.}
	\label{fig:introduction}
\end{figure}

Semantic parsing is the task of translating text to a formal meaning
representation such as logical forms or structured queries.  There has
recently been a surge of interest in developing machine learning
methods for semantic parsing (see the references in
Section~\ref{sec:related-work}), due in part to the existence of
corpora containing utterances annotated with formal meaning
representations. Figure~\ref{fig:introduction} shows an example of a
question (left hand-side) and its annotated logical form (right
hand-side), taken from \textsc{Jobs}~\cite{cocktail}, a well-known
semantic parsing benchmark. In order to predict the correct logical
form for a given utterance, most previous systems rely on predefined
templates and manually designed features, which often render the
parsing model domain- or representation-specific.  In this work, we
aim to use a simple yet effective method to bridge the gap between
natural language and logical form with minimal domain knowledge.

Encoder-decoder architectures based on recurrent neural networks
have been successfully applied to a variety of NLP tasks ranging from
syntactic parsing~\cite{grammar:foreign:language}, to machine
translation~\cite{mt:kalchbrenner13emnlp,mt:cho-EtAl:2014:EMNLP2014,mt:seq2seq},
and image description
generation~\cite{imagecaption:rnn:stanford,imagecaption:lstm:google}.
As shown in Figure~\ref{fig:introduction}, we adapt the general
encoder-decoder paradigm to the semantic parsing task. Our model
learns from natural language descriptions paired with meaning
representations; it encodes sentences and decodes logical forms using
recurrent neural networks with long short-term memory (LSTM) units.
We present two model variants, the first one treats semantic parsing
as a vanilla sequence transduction task, whereas our second model is
equipped with a hierarchical tree decoder which explicitly captures
the compositional structure of logical forms. We also introduce an
attention
mechanism~\cite{mt:jointly:align:translate,luong-pham-manning:2015:EMNLP}
allowing the model to learn soft alignments between natural language
and logical forms and present an argument identification step to
handle rare mentions of entities and numbers.

Evaluation results demonstrate that compared to previous methods our
model achieves similar or better performance across datasets and
meaning representations, despite using no hand-engineered domain- or
representation-specific features.

\section{Related Work}
\label{sec:related-work}

Our work synthesizes two strands of research, namely semantic parsing
and the encoder-decoder architecture with neural networks.

The problem of learning semantic parsers has received significant
attention, dating back to \newcite{lunar}. Many approaches learn from
sentences paired with logical forms following various modeling
strategies. Examples include the use of parsing models
\cite{miller96,scissor,lnlz08,tisp}, inductive logic programming
\cite{chill,tang-mooney:2000:EMNLP,Thomson:Mooney:2003}, probabilistic
automata \cite{He:Young:2006}, string/tree-to-tree transformation
rules \cite{silt}, classifiers based on string kernels \cite{krisp},
machine translation \cite{wasp,lambdawasp,sp:as:mt}, and combinatory
categorial grammar induction techniques \cite{zc05,zc07,ubl,fubl}.
Other work learns semantic parsers without relying on logical-from
annotations, e.g.,~from sentences paired with conversational logs
\cite{artzi-zettlemoyer:2011:EMNLP}, system demonstrations
\cite{Chen:Mooney:2011,Goldwasser:Roth:2011,Artzi:Zettlemoyer:2013},
question-answer pairs \cite{clarke2010driving,dcs}, and distant
supervision
\cite{krishnamurthy2012weakly,cai2013semantic,reddy2014semanticparsing}.

Our model learns from natural language descriptions paired with
meaning representations. Most previous systems rely on high-quality
lexicons, manually-built templates, and features which are either
domain- or representation-specific. We instead present a general
method that can be easily adapted to different domains and meaning
representations. We adopt the general encoder-decoder framework based
on neural networks which has been recently repurposed for various NLP
tasks such as syntactic parsing~\cite{grammar:foreign:language},
machine
translation~\cite{mt:kalchbrenner13emnlp,mt:cho-EtAl:2014:EMNLP2014,mt:seq2seq},
image description
generation~\cite{imagecaption:rnn:stanford,imagecaption:lstm:google},
question answering \cite{hermann2015teaching}, and summarization
\cite{rush2015neural}.

\newcite{neural:instruction} use a sequence-to-sequence model to map
navigational instructions to actions. Our model works on more
well-defined meaning representations (such as Prolog and lambda
calculus) and is conceptually simpler; it does not employ
bidirectionality or multi-level alignments. \newcite{deep:sp} propose
a different architecture for semantic parsing based on the combination
of two neural network models.  The first model learns shared
representations from pairs of questions and their translations into
knowledge base queries, whereas the second model generates the queries
conditioned on the learned representations. However, they do not
report empirical evaluation results.

\section{Problem Formulation}
\label{sec:methods}

Our aim is to learn a model which maps natural language input $q = x_1
\cdots x_{|q|} $ to a logical form representation of its meaning~$a =
y_1 \cdots y_{|a|}$.  The conditional probability~$p\left( a | q
\right)$ is decomposed as:
\begin{equation}
	\label{eq:prob:whole:seq}
	p\left( a | q \right) = \prod _{ t = 1 }^{ |a| }{ p\left( y_t | y_{<t} , q \right) }
\end{equation}
where $y_{<t} = y_1 \cdots y_{t-1}$.

Our method consists of an \textbf{encoder} which encodes natural
language input $q$ into a vector representation and a \textbf{decoder}
which learns to generate $y_1, \cdots, y_{|a|}$ conditioned on the
encoding vector.  In the following we describe two models varying in
the way in which~$p\left( a | q \right)$ is computed.

\subsection{Sequence-to-Sequence Model}
\label{sec:seq2seq-model}

\begin{figure}[t]
	\centering
	\includegraphics[width=0.42\textwidth]{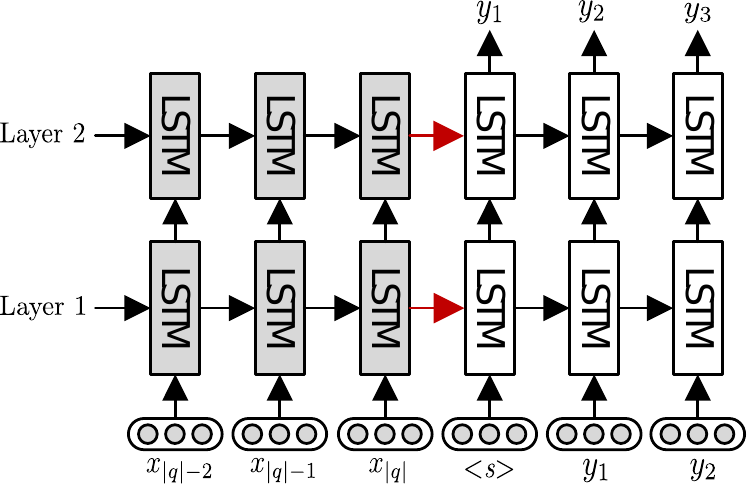}
	\caption{Sequence-to-sequence (\textsc{Seq2Seq}) model with two-layer recurrent
		neural networks.}
	\label{fig:seq2seq}
\end{figure}

This model regards both input $q$ and output $a$ as sequences.  As
shown in Figure~\ref{fig:seq2seq}, the encoder and decoder are two
different $L$-layer recurrent neural networks with long short-term
memory (LSTM) units which recursively process tokens one by one.  The
first $|q|$ time steps belong to the encoder, while the
following~$|a|$ time steps belong to the decoder.  Let ${\mathbf{h}
}_{ t }^{ l } \in \mathbb{R}^{n}$ denote the hidden vector at time
step~$t$ and layer~$l$.  ${\mathbf{h} }_{ t }^{ l }$ is then computed
by:
\begin{equation}
	{\mathbf{h} }_{ t }^{ l } = \text{LSTM} \left( {\mathbf{h} }_{ t-1 }^{ l }, {\mathbf{h} }_{ t }^{ l-1 } \right)
\end{equation}
where $\text{LSTM}$ refers to the LSTM function being used. In our experiments
we follow the architecture described
in~\newcite{zaremba2014recurrent}, however other types of gated
activation functions are possible (e.g.,
\newcite{mt:cho-EtAl:2014:EMNLP2014}).  For the encoder, ${\mathbf{h} }_{
  t }^{ 0 } = {\mathbf{W}_q} \mathbf{e}(x_t)$ is the word vector of
the current input token, with~$\mathbf{W}_q \in \mathbb{R}^{n \times
  |V_q|}$ being a parameter matrix, and $\mathbf{e}(\cdot)$ the index
of the corresponding token. For the decoder, ${\mathbf{h} }_{ t }^{ 0
} = {\mathbf{W}_a} \mathbf{e}(y_{t-1})$ is the word vector of the
previous predicted word, where $\mathbf{W}_a \in \mathbb{R}^{n \times
  |V_a|}$.
Notice that the encoder and decoder have different LSTM parameters.


Once the tokens of the input sequence $x_1, \cdots, x_{|q|}$ are
encoded into vectors, they are used to initialize the hidden states of
the first time step in the decoder.  Next, the hidden vector of the
topmost LSTM~${\mathbf{h} }_{ t }^{ L }$ in the decoder is used to
predict the \mbox{$t$-th}~output token as:
\begin{equation}
	\label{eq:seq2seq:decoder:predict}
	p\left( y_t | y_{<t} , q \right) = {\softmax \left( \mathbf{W}_o {\mathbf{h} }_{ t }^{ L } \right)}^{\intercal} \mathbf{e}\left( y_t \right)
\end{equation}
where $\mathbf{W}_o \in \mathbb{R}^{|V_a| \times n}$ is a parameter
matrix, and $\mathbf{e}\left( y_t \right) \in \{0,1\}^{|V_a|}$ a
one-hot vector for computing $y_t$'s~probability from the predicted
distribution.

We augment every sequence with a
``start-of-sequence''~\textit{\textless s\textgreater} and
``end-of-sequence'' \textit{\textless /s\textgreater}~token. The
generation process terminates once~\textit{\textless /s\textgreater}
is predicted.  The conditional probability of generating the whole
sequence $p\left( a | q \right)$ is then obtained using
Equation~\eqref{eq:prob:whole:seq}.

\subsection{Sequence-to-Tree Model}
\label{sec:seq2tree-model}

\begin{figure}[t]
	\centering
	\includegraphics[width=0.5\textwidth]{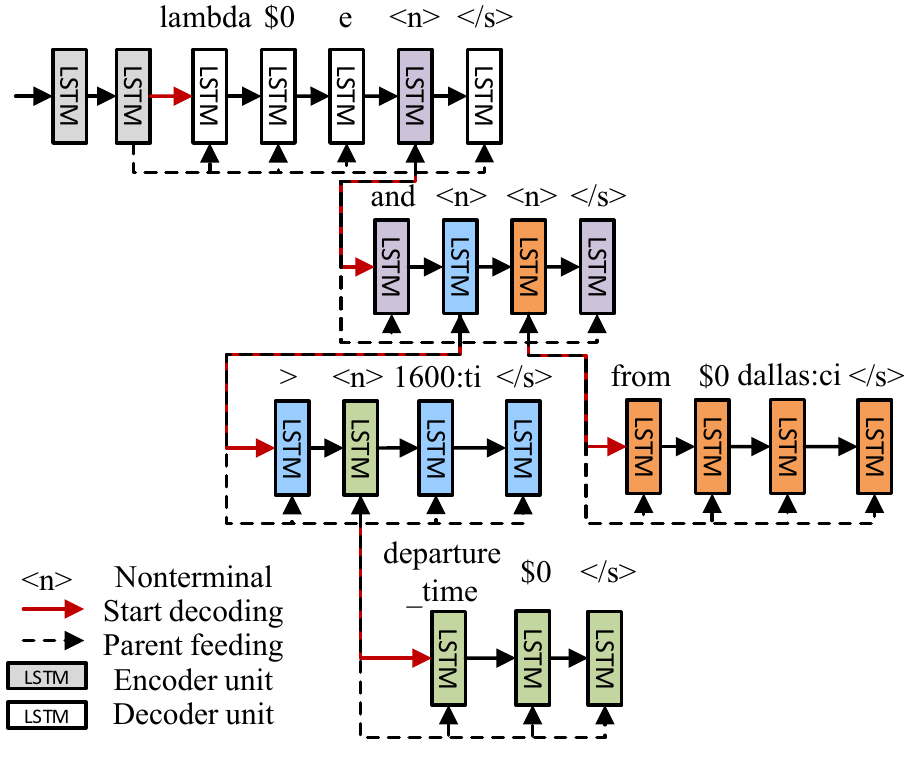}
	\caption{Sequence-to-tree (\textsc{Seq2Tree}) model with a
          hierarchical tree decoder.}
	\label{fig:seq2tree}
\end{figure}

The \textsc{Seq2Seq} model has a potential drawback in that it ignores
the hierarchical structure of logical forms. As a result, it needs to
memorize various pieces of auxiliary information (e.g., bracket pairs)
to generate well-formed output. In the following we present a
hierarchical tree decoder which is more faithful to the compositional
nature of meaning representations. A schematic description of the
model is shown in Figure~\ref{fig:seq2tree}.

The present model shares the same encoder with the
sequence-to-sequence model described in
Section~\ref{sec:seq2seq-model} (essentially it learns to encode
input~$q$ as vectors).  However, its decoder is fundamentally
different as it generates logical forms in a top-down manner.  In
order to represent tree structure, we define a ``nonterminal''
\textit{\textless n\textgreater}~token which indicates subtrees.  As
shown in Figure~\ref{fig:seq2tree}, we preprocess the logical form
``\textit{lambda \$0 e (and (\textgreater (departure\_time \$0)
  1600:ti) (from \$0 dallas:ci))}'' to a tree by replacing tokens
between pairs of brackets with nonterminals. Special tokens
\textit{\textless s\textgreater} and \textit{\textless (\textgreater}
denote the beginning of a sequence and nonterminal sequence,
respectively (omitted from Figure~\ref{fig:seq2tree} due to lack of
space). Token \textit{\textless /s\textgreater} represents the end of
sequence.

After encoding input~$q$, the hierarchical tree decoder uses recurrent
neural networks to generate tokens at depth~1 of the subtree
corresponding to parts of logical form $a$. If the predicted token
is~\textit{\textless n\textgreater}, we decode the sequence by
conditioning on the nonterminal's hidden vector. This process
terminates when no more nonterminals are emitted.  In other words, a
sequence decoder is used to hierarchically generate the tree
structure.

In contrast to the sequence decoder described in
Section~\ref{sec:seq2seq-model}, the current hidden state does not
only depend on its previous time step. In order to better utilize the
parent nonterminal's information, we introduce a \emph{parent-feeding}
connection where the hidden vector of the parent nonterminal is
concatenated with the inputs and fed into LSTM.

\begin{figure}[t]
	\centering
	\includegraphics[width=0.4\textwidth]{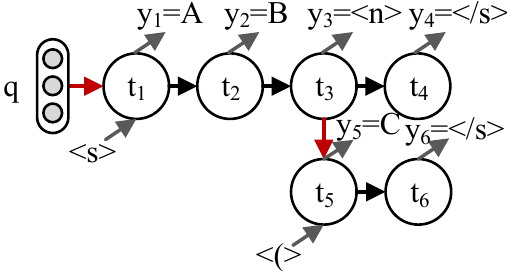}
	\caption{A \textsc{Seq2Tree} decoding example for the logical form ``\textit{A B (C)}''.}
	\label{fig:seq2tree:example}
\end{figure}

As an example,
Figure~\ref{fig:seq2tree:example} shows the decoding tree
corresponding to the logical form ``\textit{A B (C)}'', where $y_1
\cdots y_6$ are predicted tokens, and $t_1 \cdots t_6$ denote
different time steps. Span ``\textit{(C)}'' corresponds to a subtree.
Decoding in this example has two steps: once input $q$ has been
encoded, we first generate $y_1 \cdots y_4$ at depth $1$ until token
\textit{\textless /s\textgreater} is predicted; next, we generate
$y_5, y_6$ by conditioning on nonterminal $t_3$'s hidden vectors.  The
probability $p\left( a | q \right)$ is the product of these two
sequence decoding steps:
\begin{equation}
\label{eq:prob:seq2tree}
p\left( a | q \right) = p \left( y_1 y_2 y_3 y_4 | q \right) p \left( y_5 y_6 | y_{\leq 3}, q \right) 
\end{equation}
where Equation~(\ref{eq:seq2seq:decoder:predict}) is used for the
prediction of each output token.

\subsection{Attention Mechanism}
\label{sec:attention-mechanism}

As shown in Equation~\eqref{eq:seq2seq:decoder:predict}, the hidden
vectors of the input sequence are not directly used in the decoding
process. However, it makes intuitively sense to consider relevant
information from the input to better predict the current
token. Following this idea, various techniques have been proposed to
integrate encoder-side information (in the form of a context vector)
at each time step of the decoder
\cite{mt:jointly:align:translate,luong-pham-manning:2015:EMNLP,imagecaption:attend}.

\begin{figure}[t]
	\centering
	\includegraphics[width=0.42\textwidth]{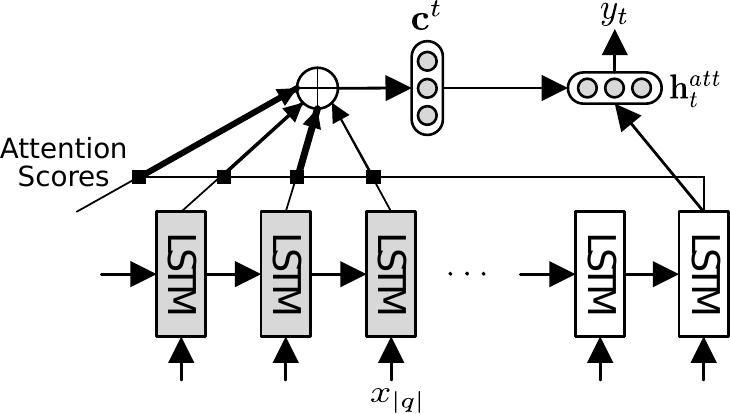}
	\caption{Attention scores are computed by the current hidden vector and all the hidden vectors of encoder. Then, the encoder-side context vector ${\mathbf{c} }^{ t }$ is obtained in the form of a weighted sum, which is further used to predict $y_t$.}
	\label{fig:attention}
\end{figure}

As shown in Figure~\ref{fig:attention}, in order to find relevant
encoder-side context for the current hidden state~${\mathbf{h} }_{ t }^{ L }$ of decoder, we compute its attention score with the
\mbox{$k$-th}~hidden state in the encoder as:
\begin{equation}
	\label{eq:attention:score}
	{ s }_{ k }^{ t } = \frac { \exp \{ {\mathbf{h} }_{ k }^{ L } \cdot {\mathbf{h} }_{ t }^{ L } \} }{ \sum _{ j=1 }^{ |q| }{ \exp \{ {\mathbf{h} }_{ j }^{ L } \cdot {\mathbf{h} }_{ t }^{ L } \} }  } 
\end{equation}
where ${\mathbf{h} }_{ 1 }^{ L }, \cdots, {\mathbf{h} }_{ |q| }^{ L }$ are the top-layer hidden vectors of the encoder.
Then, the context vector is the weighted sum of the hidden vectors in the encoder:
\begin{equation}
	\mathbf{c}^{t} = \sum_{ k=1 }^{ |q| }{ { s }_{ k }^{ t } {\mathbf{h} }_{ k }^{ L } }
\end{equation}

In lieu of Equation~\eqref{eq:seq2seq:decoder:predict}, we further use
this context vector which acts as a summary of the encoder to compute
the probability of generating $y_t$ as:
\begin{equation}
	\label{eq:attention:new:hidden}
	{\mathbf{h} }_{ t }^{ att } = \tanh \left( \mathbf{W}_1 {\mathbf{h} }_{ t }^{ L } + \mathbf{W}_2 \mathbf{c}^{t} \right)
\end{equation}
\begin{equation}
	\label{eq:attention:decoder:predict}
	p\left( y_t | y_{<t} , q \right) = {\softmax \left( \mathbf{W}_o {\mathbf{h} }_{ t }^{ att } \right)}^{\intercal} \mathbf{e}\left( y_t \right)
\end{equation}
where $\mathbf{W}_o \in \mathbb{R}^{|V_a| \times n}$ and $\mathbf{W}_1 , \mathbf{W}_2 \in
\mathbb{R}^{n \times n}$ are three parameter matrices, and
$\mathbf{e}\left( y_t \right)$ is a one-hot vector used to obtain
$y_t$'s~probability.

\subsection{Model Training}
\label{sec:model-training}

Our goal is to maximize the likelihood of the generated logical forms
given natural language utterances as input. So the objective function
is:
\begin{equation}
	\label{eq:objective}
	\minimize -\sum_{(q , a) \in \mathcal{D} }{ \log{p \left( a | q \right)}}
\end{equation}
where $\mathcal{D}$ is the set of all natural language-logical form
training pairs, and~$p \left( a | q \right)$ is computed as shown in
Equation~\eqref{eq:prob:whole:seq}.

The RMSProp algorithm~\cite{rmsprop} is employed to solve this
non-convex optimization problem.  Moreover, dropout is used for
regularizing the model \cite{zaremba2014recurrent}. Specifically,
dropout operators are used between different LSTM layers and for the
hidden layers before the softmax classifiers. This technique can
substantially reduce overfitting, especially on datasets of small
size.

\subsection{Inference}
\label{sec:inference}

At test time, we predict the logical form for an input
utterance~$q$ by:
\begin{equation}
	\label{eq:inference}
	\hat{a} = \argmax_{a'}{ p \left( a' | q \right) }
\end{equation}
where~$a'$ represents a candidate output.  However, it
is impractical to iterate over all possible results to obtain the
optimal prediction.  According to Equation~\eqref{eq:prob:whole:seq},
we decompose the probability~$p\left( a | q \right)$ so that we can
use greedy search (or beam search) to generate tokens one by one.

Algorithm~1 describes the decoding process for \textsc{Seq2Tree}.  The
time complexity of both decoders is $\mathcal{O}(|a|)$, where $|a|$ is the
length of output. The extra computation of \textsc{Seq2Tree} compared
with \textsc{Seq2Seq} is to maintain the nonterminal queue, which can
be ignored because most of time is spent on matrix operations. We
implement the hierarchical tree decoder in a batch mode, so that it
can fully utilize GPUs. Specifically, as shown in Algorithm~1, every
time we pop multiple nonterminals from the queue and decode these
nonterminals in one batch.

%
\begin{figure}[t]
\centering
\includegraphics[width=.48\textwidth]{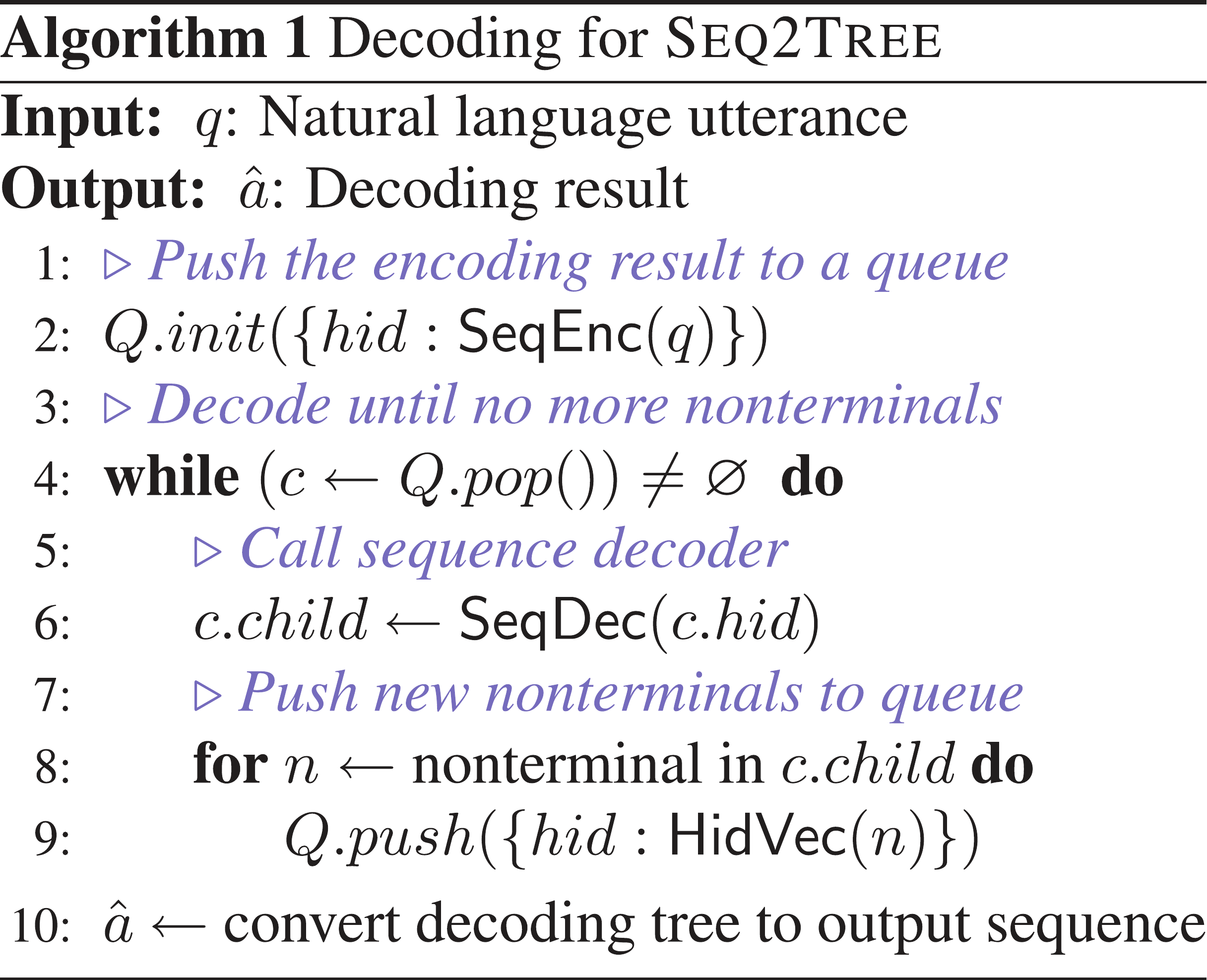}
\end{figure}

\subsection{Argument Identification}
\label{sec:argument-identification}

The majority of semantic parsing datasets have been developed with
question-answering in mind. In the typical application setting,
natural language questions are mapped into logical forms and executed
on a knowledge base to obtain an answer. Due to the nature of the
question-answering task, many natural language utterances contain
entities or numbers that are often parsed as arguments in the logical
form. Some of them are unavoidably rare or do not appear in the
training set at all (this is especially true for small-scale
datasets).  Conventional sequence encoders simply replace rare words
with a special unknown word
symbol~\cite{mt:rare:word:google,mt:rare:word:cho}, which would be
detrimental for semantic parsing.

We have developed a simple procedure for argument
identification. Specifically, we identify entities and numbers in
input questions and replace them with their type names and unique IDs.
For instance, we pre-process the training example ``\textit{jobs with
  a salary of 40000}'' and its logical form ``\textit{job(ANS),
salary\_greater\_than(ANS, 40000, year)}'' as ``\textit{jobs with a
  salary of \enttype{num}{0}}'' and ``\textit{job(ANS),
salary\_greater\_than(ANS, \enttype{num}{0}, year)}''.  We use the
pre-processed examples as training data.  At inference time, we also
mask entities and numbers with their types and IDs. Once we obtain the
decoding result, a post-processing step recovers all the markers
\enttype{type}{i} to their corresponding logical constants.

\section{Experiments}
\label{sec:experiments}

We compare our method against multiple previous systems on four
datasets.  We describe these datasets below, and present our
experimental settings and results. Finally, we conduct model
analysis in order to understand what the model learns.
The code is available at
\url{https://github.com/donglixp/lang2logic}.

\subsection{Datasets}
\label{sec:datasets}

\begin{table*}[t]
	\centering
	\small
	\begin{tabular}{l c l}
		\hline
		\textbf{Dataset} & \textbf{Length} & \textbf{Example} \\ \hline
		\textsc{Jobs} & \tabincell{r}{9.80 \\ 22.90} & \tabincell{l}{\textit{what microsoft jobs do not require a bscs?} \\ {answer(company(J,'microsoft'),job(J),not((req\_deg(J,'bscs'))))}} \\ \hline
		\textsc{Geo} & \tabincell{r}{7.60 \\ 19.10} & \tabincell{l}{\textit{what is the population of the state with the largest area?} \\ {(population:i (argmax \$0 (state:t \$0) (area:i \$0)))}} \\ \hline
		\textsc{Atis} & \tabincell{r}{11.10 \\ 28.10} & \tabincell{l}{\textit{dallas to san francisco leaving after 4 in the afternoon please} \\{\small {(lambda \$0 e (and (\textgreater (departure\_time \$0) 1600:ti) (from \$0 dallas:ci) (to \$0 san\_francisco:ci)))}}} \\ \hline
		\textsc{Ifttt} & \tabincell{r}{6.95 \\ 21.80} & \tabincell{l}{\textit{Turn on heater when temperature drops below 58 degree} \\ {\footnotesize {TRIGGER: Weather - Current\_temperature\_drops\_below - ((Temperature (58)) (Degrees\_in (f))) } } \\ {\footnotesize {ACTION: WeMo\_Insight\_Switch - Turn\_on - ((Which\_switch? ("")))} }} \\ \hline
	\end{tabular}
	\normalsize
	\caption{Examples of natural language descriptions and their meaning representations from four datasets. The average length of input and output sequences is shown in the second column.}
	\label{table:dataset}
\end{table*}

Our model was trained on the following datasets, covering different
domains and using different meaning representations.  Examples for
each domain are shown in Table~\ref{table:dataset}.

\paragraph{\textsc{Jobs}} This benchmark dataset contains~$640$
queries to a database of job listings.  Specifically, questions are
paired with Prolog-style queries. We used the same
training-test split as~\newcite{zc05} which contains $500$~training
and $140$~test instances.  Values for the variables company, degree,
language, platform, location, job area, and number are identified.

\noindent
\paragraph{\textsc{Geo}} This is a standard semantic parsing benchmark
which contains~$880$ queries to a database of
U.S. geography. \textsc{Geo} has $880$ instances split
into a training set of $680$ training examples and $200$ test
examples~\cite{zc05}. We used the same meaning representation based on
lambda-calculus as~\newcite{fubl}.  Values for the variables city,
state, country, river, and number are identified.

\noindent
\paragraph{\textsc{Atis}} This dataset has $5,410$ queries to a flight
booking system. The standard split has $4,480$ training
instances, $480$ development instances, and $450$ test
instances. Sentences are paired with lambda-calculus expressions.
Values for the variables date, time, city, aircraft code, airport,
airline, and number are identified.

\noindent
\paragraph{\textsc{Ifttt}} \newcite{ifttt} created this dataset by
extracting a large number of if-this-then-that recipes from the \textsc{Ifttt}
website\footnote{\url{http://www.ifttt.com}}. Recipes are simple
programs with exactly one trigger and one action which users specify
on the site. Whenever the conditions of the trigger are satisfied, the
action is performed. Actions typically revolve around home security
(e.g.,~``\textit{turn on my lights when I arrive home}''), automation
(e.g.,~``\textit{text me if the door opens}''), well-being
(e.g.,~``\textit{remind me to drink water if I've been at a bar for
	more than two hours}''), and so on.  Triggers and actions are
selected from different channels ($160$ in total) representing various
types of services, devices (e.g., Android), and knowledge sources
(such as ESPN or Gmail).  In the dataset, there are~$552$~trigger
functions from~$128$ channels, and~$229$ action functions from~$99$
channels.  We used Quirk et al.'s \shortcite{ifttt} original split
which contains~$77,495$ training, $5,171$~development, and
$4,294$~test examples.  The \textsc{Ifttt} programs are represented as
abstract syntax trees and are paired with natural language
descriptions provided by users (see Table~\ref{table:dataset}).
Here, numbers and URLs are identified.

\subsection{Settings}
\label{sec:settings}

Natural language sentences were lowercased; misspellings were
corrected using a dictionary based on the Wikipedia list of common
misspellings. Words were stemmed using
NLTK~\cite{nltk}. For \textsc{Ifttt}, we
filtered tokens, channels and functions which appeared less than five
times in the training set. For the other datasets, we filtered input
words which did not occur at least two times in the training set, but
kept all tokens in the logical forms.
Plain string matching was employed to identify augments as described in Section~\ref{sec:argument-identification}. More sophisticated approaches could be used, however we leave this future work.

Model hyper-parameters were cross-validated on the training set for
\textsc{Jobs} and \textsc{Geo}. We used the standard development sets
for \textsc{Atis} and \textsc{Ifttt}.
We used the RMSProp algorithm (with batch size set to~$20$) to update
the parameters.  The smoothing constant of RMSProp was~$0.95$.
Gradients were clipped at $5$ to alleviate the exploding gradient
problem~\cite{pascanu2013difficulty}.  Parameters were randomly
initialized from a uniform distribution
$\mathcal{U}\left(-0.08,0.08\right)$.  A two-layer LSTM was used for
\textsc{Ifttt}, while a one-layer LSTM was employed for the other
domains.  The dropout rate was selected from
$\{0.2,0.3,0.4,0.5\}$. Dimensions of hidden vector and word embedding
were selected from $\{150,200,250\}$.  Early stopping was used to
determine the number of epochs.  Input sentences were reversed before
feeding into the encoder~\cite{mt:seq2seq}.  We use greedy search to
generate logical forms during inference.  Notice that two decoders
with shared word embeddings were used to predict triggers and actions
for \textsc{Ifttt}, and two softmax classifiers are used to classify
channels and functions.

\subsection{Results}
\label{sec:results}

\begin{table}[t]
	\centering
	\small
	\begin{tabular}{l c}
		\hline
		\textbf{Method}          & \textbf{Accuracy} \\ \hline
		COCKTAIL~\cite{cocktail} & 79.4              \\
		PRECISE~\cite{precise}   & 88.0              \\
		ZC05~\cite{zc05}         & 79.3              \\
		DCS+L~\cite{dcs}         & 90.7              \\
		TISP~\cite{tisp}         & 85.0              \\ \hline
		\textsc{Seq2Seq}         & 87.1              \\
		~~~~$-$ attention        & 77.9              \\
		~~~~$-$ argument         & 70.7              \\
		\textsc{Seq2Tree}        & 90.0              \\
		~~~~$-$ attention        & 83.6              \\ \hline
	\end{tabular}
	\normalsize
	\caption{Evaluation results on \textsc{Jobs}.}
	\label{table:results:jobs}
\end{table}

\begin{table}[t]
	\centering
	\small
	\begin{tabular}{l c}
		\hline
		\textbf{Method}                  & \textbf{Accuracy} \\ \hline
		SCISSOR~\cite{scissor}           & 72.3              \\
		KRISP~\cite{krisp}               & 71.7              \\
		WASP~\cite{wasp}                 & 74.8              \\
		$\lambda$-WASP~\cite{lambdawasp} & 86.6              \\
		LNLZ08~\cite{lnlz08}             & 81.8              \\ \hline
		ZC05~\cite{zc05}                 & 79.3              \\
		ZC07~\cite{zc07}                 & 86.1              \\
		UBL~\cite{ubl}                   & 87.9              \\
		FUBL~\cite{fubl}                 & 88.6              \\
		KCAZ13~\cite{onthefly13}         & 89.0              \\
		DCS+L~\cite{dcs}                 & 87.9              \\
		TISP~\cite{tisp}                 & 88.9              \\ \hline
		\textsc{Seq2Seq}                 & 84.6              \\
		~~~~$-$ attention                & 72.9              \\
		~~~~$-$ argument                 & 68.6              \\
		\textsc{Seq2Tree}                & 87.1              \\
		~~~~$-$ attention                & 76.8              \\ \hline
	\end{tabular}
	\normalsize
	\caption{Evaluation results on \textsc{Geo}. 10-fold
		cross-validation is used for the systems shown in the top
		half of the table. The standard split of ZC05 is used
		for all other systems.}
	\label{table:results:GeoQueries}
\end{table}

\begin{table}[t]
	\centering
	\small
	\begin{tabular}{l c}
		\hline
		\textbf{Method}       & \textbf{Accuracy} \\ \hline
		ZC07~\cite{zc07}      & 84.6              \\
		UBL~\cite{ubl}        & 71.4              \\
		FUBL~\cite{fubl}      & 82.8              \\
		GUSP-FULL~\cite{gusp} & 74.8              \\
		GUSP++~\cite{gusp}    & 83.5              \\
		TISP~\cite{tisp}      & 84.2              \\ \hline
		\textsc{Seq2Seq}      & 84.2              \\
		~~~~$-$ attention     & 75.7              \\
		~~~~$-$ argument      & 72.3              \\
		\textsc{Seq2Tree}     & 84.6              \\
		~~~~$-$ attention     & 77.5              \\ \hline
	\end{tabular}
	\normalsize
	\caption{Evaluation results on \textsc{Atis}.}
	\label{table:results:ATIS}
\end{table}

\begin{table}[t]
	\centering
	\small
	\subfloat[Omit non-English. \label{table:results:IFTTT:1}] {%
		\begin{tabular}{l c c c}
			\hline
			\textbf{Method}   & \textbf{Channel} & \textbf{+Func} & \textbf{F1} \\ \hline
			retrieval         & 28.9             & 20.2           & 41.7        \\
			phrasal           & 19.3             & 11.3           & 35.3        \\
			sync              & 18.1             & 10.6           & 35.1        \\
			classifier        & 48.8             & 35.2           & 48.4        \\
			posclass          & 50.0             & 36.9           & 49.3        \\ \hline
			\textsc{Seq2Seq}  & 54.3             & 39.2           & 50.1        \\
			~~~~$-$ attention & 54.0             & 37.9           & 49.8        \\
			~~~~$-$ argument  & 53.9             & 38.6           & 49.7        \\
			\textsc{Seq2Tree} & 55.2             & 40.1           & 50.4        \\
			~~~~$-$ attention & 54.3             & 38.2           & 50.0        \\ \hline
		\end{tabular}
	}\\%
	\subfloat[Omit non-English \& unintelligible.
	\label{table:results:IFTTT:2}]{%
		\begin{tabular}{l c c c}
			\hline
			\textbf{Method}   & \textbf{Channel} & \textbf{+Func} & \textbf{F1} \\ \hline
			retrieval         & 36.8             & 25.4           & 49.0        \\
			phrasal           & 27.8             & 16.4           & 39.9        \\
			sync              & 26.7             & 15.5           & 37.6        \\
			classifier        & 64.8             & 47.2           & 56.5        \\
			posclass          & 67.2             & 50.4           & 57.7        \\ \hline
			\textsc{Seq2Seq}  & 68.8             & 50.5           & 60.3        \\
			~~~~$-$ attention & 68.7             & 48.9           & 59.5        \\
			~~~~$-$ argument  & 68.8             & 50.4           & 59.7        \\
			\textsc{Seq2Tree} & 69.6             & 51.4           & 60.4        \\
			~~~~$-$ attention & 68.7             & 49.5           & 60.2        \\ \hline
		\end{tabular}
	}\\%
	\subfloat[$\ge 3$ turkers agree with gold.
	\label{table:results:IFTTT:3}]{%
		\begin{tabular}{l c c c}
			\hline
			\textbf{Method}   & \textbf{Channel} & \textbf{+Func} & \textbf{F1} \\ \hline
			retrieval         & 43.3             & 32.3           & 56.2        \\
			phrasal           & 37.2             & 23.5           & 45.5        \\
			sync              & 36.5             & 24.1           & 42.8        \\
			classifier        & 79.3             & 66.2           & 65.0        \\
			posclass          & 81.4             & 71.0           & 66.5        \\ \hline
			\textsc{Seq2Seq}  & 87.8             & 75.2           & 73.7        \\
			~~~~$-$ attention & 88.3             & 73.8           & 72.9        \\
			~~~~$-$ argument  & 86.8             & 74.9           & 70.8        \\
			\textsc{Seq2Tree} & 89.7             & 78.4           & 74.2        \\
			~~~~$-$ attention & 87.6             & 74.9           & 73.5        \\ \hline
		\end{tabular}
	}
	\normalsize
	\caption{Evaluation results on \textsc{Ifttt}.}
	\label{table:results:IFTTT}
\end{table}

We first discuss the performance of our model on \textsc{Jobs},
\textsc{Geo}, and \textsc{Atis}, and then examine our results on
\textsc{Ifttt}. Tables~\ref{table:results:jobs}--\ref{table:results:ATIS}
present comparisons against a variety of systems previously described
in the literature. We report results with the full models
(\textsc{Seq2Seq}, \textsc{Seq2Tree}) and two ablation variants,
i.e.,~without an attention mechanism ($-$attention) and without
argument identification ($-$argument).  We report accuracy which is
defined as the proportion of the input sentences that are correctly parsed
to their gold standard logical forms.  Notice that DCS+L, KCAZ13 and
GUSP output answers directly, so accuracy in this setting is defined
as the percentage of correct answers.

\begin{figure*}[t]
	\centering
	\subfloat[c][which jobs pay num0 that do\\ \hspace*{.44cm}not require a degid0\label{fig:exp:attention:a}]{\includegraphics[scale=0.44]{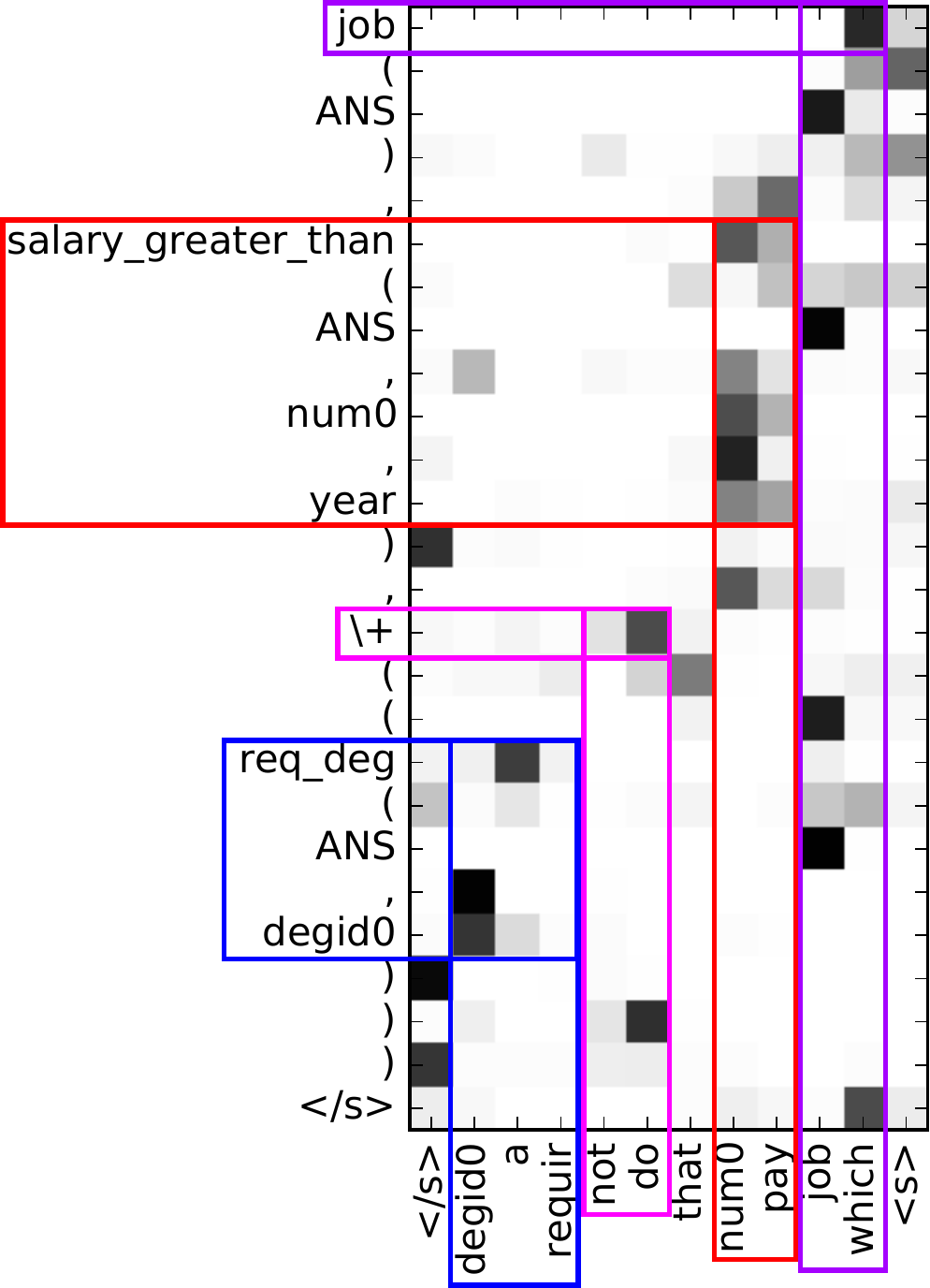}}
	\subfloat[c][what's first class fare \\ \hspace*{.44cm}round
        trip from ci0 to ci1\label{fig:exp:attention:b}]{\includegraphics[scale=0.44]{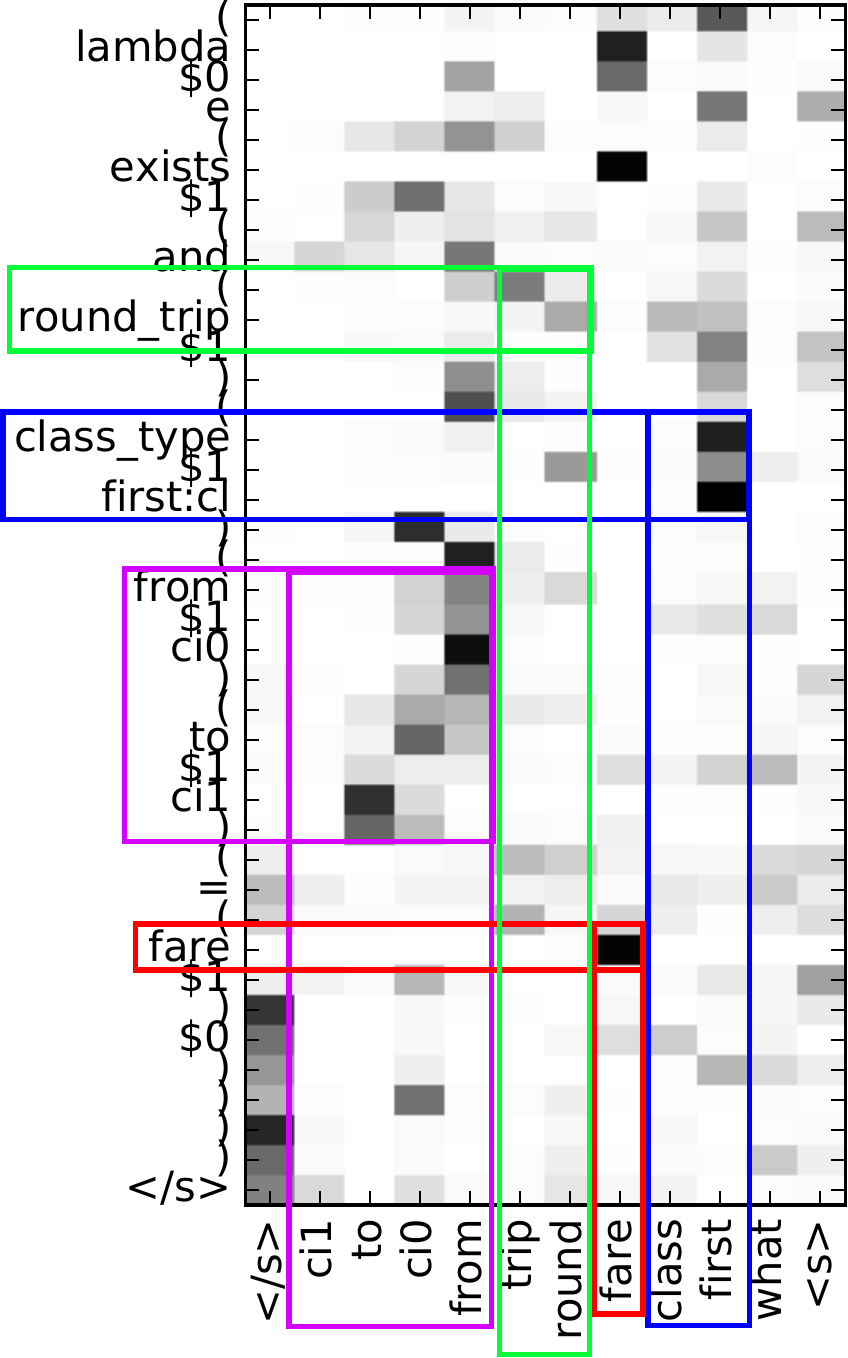}}
	\subfloat[c][what is the earliest flight \\
        \hspace*{.44cm}from ci0 to ci1
        tomorrow\label{fig:exp:attention:c}]{\includegraphics[scale=.44]{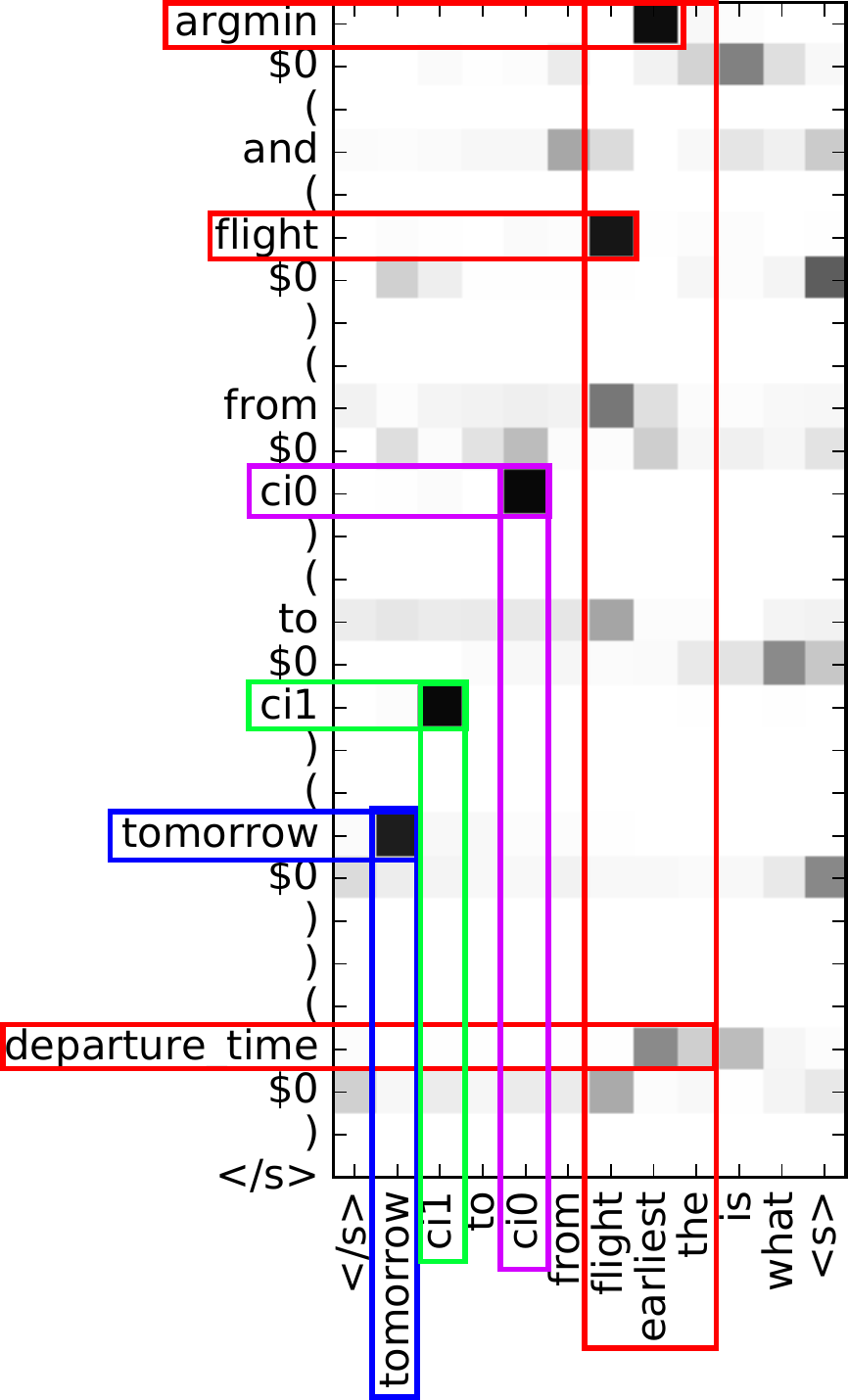}}
	\subfloat[][what is the highest elevation \hspace*{.46cm}in
        the co0\label{fig:exp:attention:d}]{\includegraphics[scale=.44]{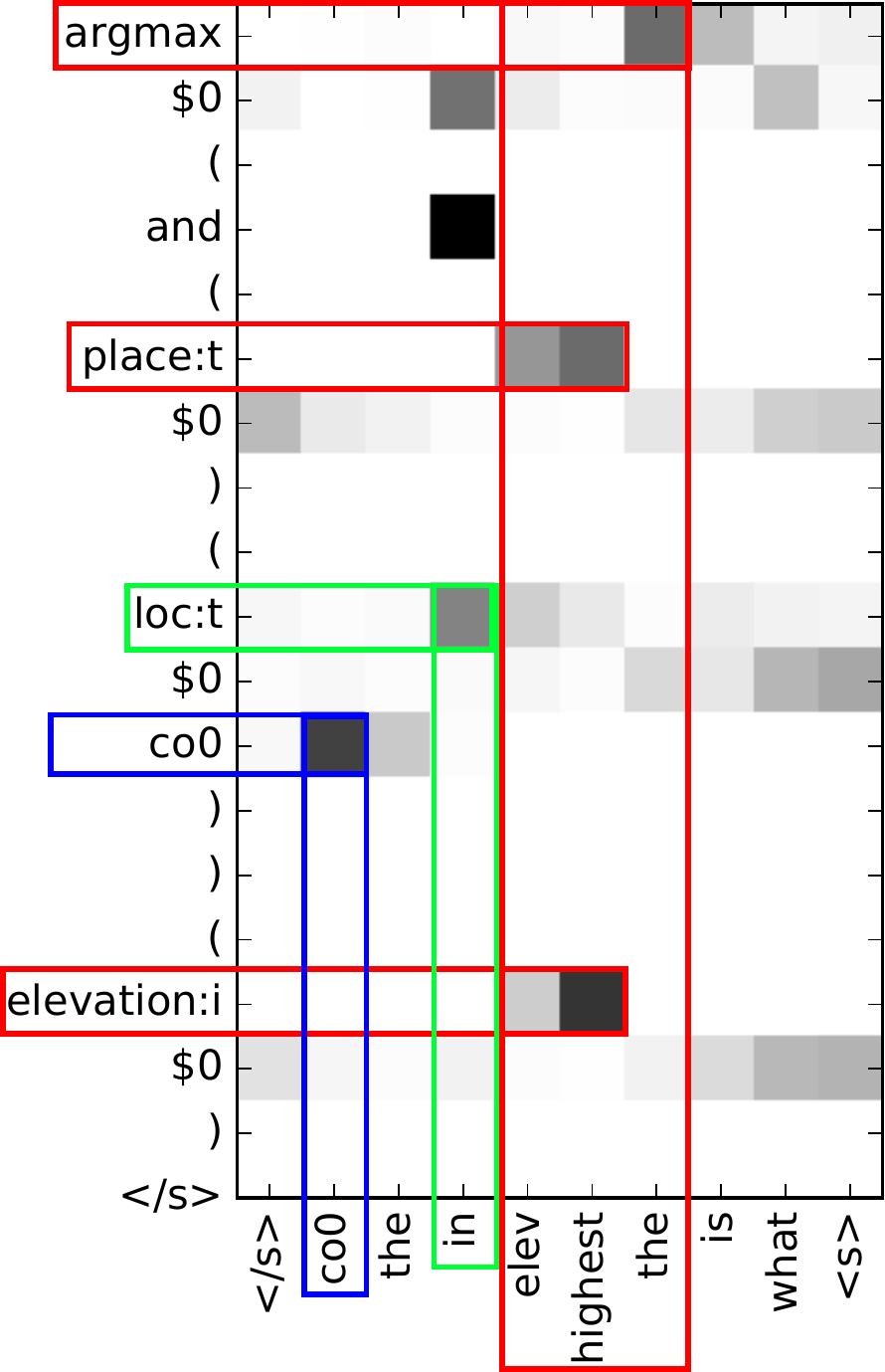}}
	\caption{Alignments (same color rectangles) produced by the
          attention mechanism (darker color represents higher
          attention score). Input sentences are reversed and
          stemmed. Model output is shown for \textsc{Seq2Seq} (a, b) and
          \textsc{Seq2Tree} (c, d). }
	\label{fig:exp:attention}
\end{figure*}

Overall, \textsc{Seq2Tree} is superior to \textsc{Seq2Seq}. This is to
be expected since \textsc{Seq2Tree} explicitly models compositional
structure. On the \textsc{Jobs} and \textsc{Geo} datasets which
contain logical forms with nested structures, \textsc{Seq2Tree}
outperforms \textsc{Seq2Seq} by~2.9\% and~2.5\%,
respectively. \textsc{Seq2Tree} achieves better accuracy over
\textsc{Seq2Seq} on \textsc{Atis} too, however, the difference is
smaller, since \textsc{Atis} is a simpler domain without complex
nested structures.  We find that adding attention substantially
improves performance on all three datasets. This underlines the
importance of utilizing soft alignments between inputs and outputs. We
further analyze what the attention layer learns in
Figure~\ref{fig:exp:attention}.  Moreover, our results show that argument
identification is critical for small-scale datasets. For example,
about $92\%$ of city names appear less than $4$ times in the
\textsc{Geo} training set, so it is difficult to learn reliable
parameters for these words.  In relation to previous work, the
proposed models achieve comparable or better performance.
Importantly, we use the same framework (\textsc{Seq2Seq} or
\textsc{Seq2Tree}) across datasets and meaning representations
(Prolog-style logical forms in \textsc{Jobs} and lambda calculus in
the other two datasets) without modification. Despite this relatively
simple approach, we observe that \textsc{Seq2Tree} ranks second on
\textsc{Jobs}, and is tied for first place with ZC07 on \textsc{Atis}.

We illustrate examples of alignments produced by \textsc{Seq2Seq} in
Figures~\ref{fig:exp:attention:a}
and~\ref{fig:exp:attention:b}. Alignments produced by
\textsc{Seq2Tree} are shown in Figures~\ref{fig:exp:attention:c}
and~\ref{fig:exp:attention:d}. Matrices of attention scores are
computed using Equation~\eqref{eq:attention:score} and are represented
in grayscale. Aligned input words and logical form predicates are
enclosed in (same color) rectangles whose overlapping areas contain
the attention scores.
Also notice that attention scores are computed by LSTM hidden vectors
which encode context information rather than just the words in their
current positions.  The examples demonstrate that the attention
mechanism can successfully model the correspondence between sentences
and logical forms, capturing reordering
(Figure~\ref{fig:exp:attention:b}), many-to-many
(Figure~\ref{fig:exp:attention:a}), and many-to-one alignments
(Figures~\ref{fig:exp:attention:c},d).

For \textsc{Ifttt}, we follow the same evaluation protocol introduced
in~\newcite{ifttt}. The dataset is extremely noisy and measuring
accuracy is problematic since predicted abstract syntax trees (ASTs)
almost never exactly match the gold standard. Quirk et al. view an AST
as a set of productions and compute balanced F1 instead which we also
adopt. The first column in Table~\ref{table:results:IFTTT} shows the
percentage of channels selected correctly for both triggers and
actions. The second column measures accuracy for \emph{both} channels
and functions. The last column shows balanced F1 against the gold tree
over all productions in the proposed derivation.  We compare our model
against posclass, the method introduced in Quirk et al. and several of
their baselines. posclass is reminiscent of KRISP~\cite{krisp}, it
learns distributions over productions given input sentences
represented as a bag of linguistic features. The retrieval baseline
finds the closest description in the training data based on
character string-edit-distance and returns the recipe for that
training program.  The phrasal method uses phrase-based machine
translation to generate the recipe, whereas sync extracts synchronous
grammar rules from the data, essentially recreating
WASP~\cite{wasp}. Finally, they use a binary classifier to predict
whether a production should be present in the derivation tree
corresponding to the description.

\newcite{ifttt} report results on the full test data and smaller
subsets after noise filtering, e.g.,~when non-English and
unintelligible descriptions are removed
(Tables~\ref{table:results:IFTTT:1}
and~\ref{table:results:IFTTT:2}). They also ran their system on a
high-quality subset of description-program pairs which were found in
the gold standard and at least three humans managed to independently
reproduce (Table~\ref{table:results:IFTTT:3}). Across all subsets our
models outperforms posclass and related baselines.  Again we observe
that \textsc{Seq2Tree} consistently outperforms \textsc{Seq2Seq},
albeit with a small margin.  Compared to the previous datasets, the
attention mechanism and our argument identification method yield less
of an improvement. This may be due to the size of \newcite{ifttt} and
the way it was created -- user curated descriptions are often of low
quality, and thus align very loosely to their corresponding ASTs.

\subsection{Error Analysis}

Finally, we inspected the output of our model in order to identify the
most common causes of errors which we summarize below.

\paragraph{Under-Mapping} The attention model used in our experiments
does not take the alignment history into consideration. So, some
question words, expecially in longer questions, may be ignored in the
decoding process.  This is a common problem for encoder-decoder models
and can be addressed by explicitly modelling the decoding coverage of
the source words
\cite{mt:coverage:2016:ACL,mt:structalignment:2016:naacl}. Keeping
track of the attention history would help adjust future attention and
guide the decoder towards untranslated source words.

\paragraph{Argument Identification}
Some mentions are incorrectly identified as arguments. For example,
the word \textsl{may} is sometimes identified as a month when it is
simply a modal verb.  Moreover, some argument mentions are
ambiguous. For instance, \textsl{6 o'clock} can be used to express
either \textsl{6 am} or \textsl{6 pm}.  We could disambiguate
arguments based on contextual information. The execution results of
logical forms could also help prune unreasonable arguments.

\paragraph{Rare Words}
Because the data size of \textsc{Jobs}, \textsc{Geo}, and
\textsc{Atis} is relatively small, some question words are rare in the
training set, which makes it hard to estimate reliable parameters for
them. One solution would be to learn word embeddings on unannotated
text data, and then use these as pretrained vectors for question
words.

\section{Conclusions}
\label{sec:conclusions}

In this paper we presented an encoder-decoder neural network model for
mapping natural language descriptions to their meaning
representations. We encode natural language utterances into vectors
and generate their corresponding logical forms as sequences or trees
using recurrent neural networks with long short-term memory
units.  Experimental results show that enhancing the model with a
hierarchical tree decoder and an attention mechanism improves
performance across the board.
Extensive comparisons with previous methods show that our approach
performs competitively, without recourse to domain- or
representation-specific features.
Directions for future work are many and varied.  For example, it would
be interesting to learn a model from question-answer pairs without
access to target logical forms.  Beyond semantic parsing, we would
also like to apply our \textsc{Seq2Tree} model to related structured
prediction tasks such as constituency parsing.

\paragraph{Acknowledgments}
We would like to thank Luke Zettlemoyer and Tom Kwiatkowski for
sharing the ATIS dataset.  The support of the European Research
Council under award number 681760 ``Translating Multiple Modalities
into Text'' is gratefully acknowledged.

\bibliography{lang2logic}
\bibliographystyle{acl2016}

\end{document}